\documentclass[10pt,twocolumn,letterpaper]{article}
\usepackage[accsupp]{axessibility}  
\usepackage{wacv}
\usepackage{times}
\usepackage{epsfig}
\usepackage{graphicx}
\usepackage{amsmath}
\usepackage{amssymb}
\usepackage{verbatim}
\usepackage[ruled,vlined,linesnumbered]{algorithm2e}
\usepackage{multirow}
\usepackage{float}
\usepackage{bbm}
\usepackage{algorithm2e}

\newtheorem{theorem}{Theorem}[section]

\newtheorem{lemma}[theorem]{Lemma}

\DeclareMathOperator*{\argmax}{arg\,max}
\DeclareMathOperator*{\argmin}{arg\,min}

\def\wacvPaperID{202} 

\wacvfinalcopy 

\ifwacvfinal
\fi

\ifwacvfinal
\usepackage[breaklinks=true,bookmarks=false]{hyperref}
\else
\usepackage[pagebackref=true,breaklinks=true,colorlinks,bookmarks=false]{hyperref}
\fi

\ifwacvfinal
\else
\fi

\begin{document}

\title{Attack Agnostic Detection of Adversarial Examples via \\ Random Subspace Analysis}

\author{Nathan Drenkow,  Neil Fendley,  Philippe Burlina  \\
The Johns Hopkins University Applied Physics Laboratory\\
Laurel, MD 20723, USA \\
\texttt{\{first.last\}@jhuapl.edu} \\
}


\maketitle
\thispagestyle{empty}

\begin{abstract}
Whilst adversarial attack detection has received considerable attention, it remains a fundamentally challenging problem from two perspectives. First, while threat models can be well-defined, attacker strategies may still vary widely within those constraints. Therefore, detection should be considered as an open-set problem, standing in contrast to most current detection approaches.  These methods take a closed-set view and train binary detectors, thus biasing detection toward attacks seen during detector training. Second, limited information is available at test time and typically confounded by nuisance factors including the label and underlying content of the image. We address these challenges via a novel strategy based on random subspace analysis. We present a technique that utilizes  properties of random projections to characterize the behavior of clean and adversarial examples across a diverse set of subspaces. The self-consistency (or inconsistency) of model activations is leveraged to discern clean from adversarial examples. Performance evaluations demonstrate that our technique ($AUC\in[0.92, 0.98]$) outperforms competing detection strategies ($AUC\in[0.30,0.79]$), while remaining truly agnostic to the attack strategy (for both targeted/untargeted attacks). It also requires significantly less calibration data (composed only of clean examples) than competing approaches to achieve this performance. 
\vspace{-2mm}
\end{abstract}

\section{Introduction}
The rise of deep learning has led to state-of-the-art advances in machine learning (ML) across almost every conceivable application.  With deep neural networks (DNNs) as the core computational elements in increasingly complex ML systems, there is greater demand for increasing DNN robustness to adversarial attacks.  

\begin{figure}[t]
\begin{center}
        \includegraphics[width=.7\linewidth]{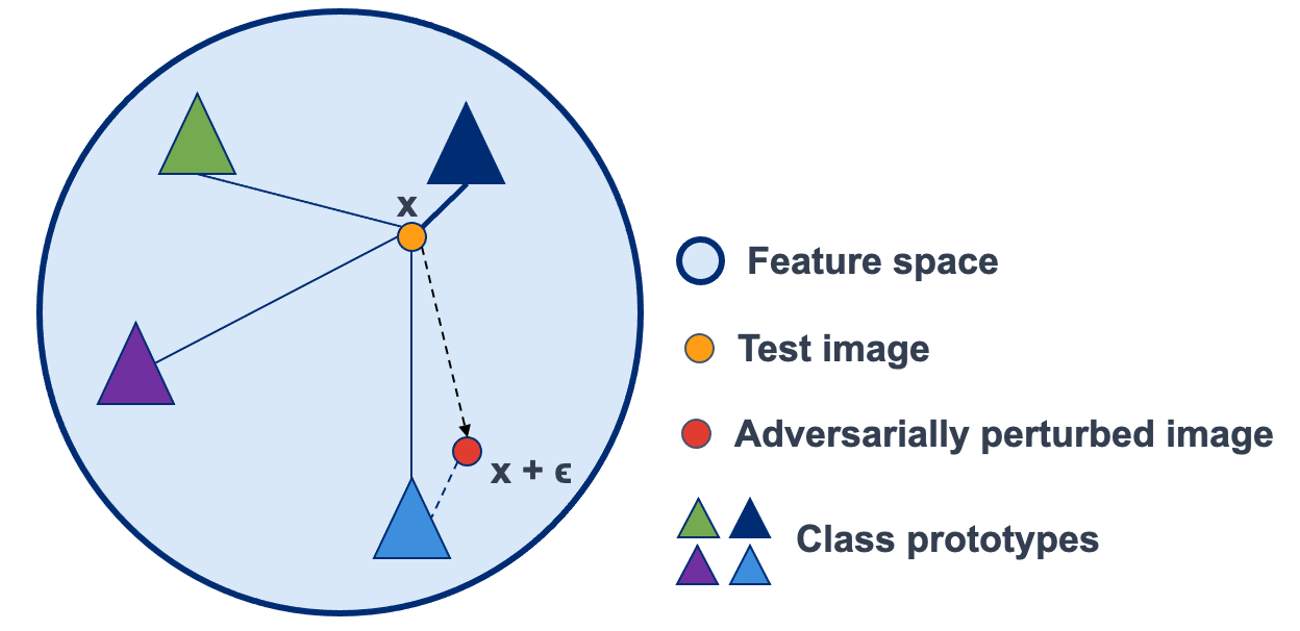} 
\end{center}
  \caption{\textit{Distances in feature space between an image embedding ($x$) and class prototypes are used as a proxy for predicting class labels.  Adversarial perturbations ($\epsilon$) shift the clean representation such that the distance to the true class prototype increases and the distance to the incorrect prototype decreases.}}
\label{fig:feature_space}
\vspace{-4mm}
\end{figure}

The discovery of adversarial examples~\cite{goodfellow2014explaining} has led to a wide range of subsequent findings in domains such as image classification, object detection, natural language processing, speech processing, and reinforcement learning.   While adversarial examples are realized in the input domain, their intended and ultimate effect is on model behavior.  To mitigate against potential attacks, we develop a model-centric, attack-agnostic method for adversarial example (AE) detection which analyzes the consistency of model activations under random projections. Our detection method is successful for a variety of attack strategies (e.g., FGSM~\cite{goodfellow2014explaining}, PGD~\cite{madry2017towards}, CWL~\cite{carlini2017adversarial}) and targeted/untargeted attacks. Our approach requires a small amount of calibration data and no \emph{a priori} attack-specific knowledge or data.

\begin{figure*}[h]
\begin{center}
    \begin{tabular}{c}
        \includegraphics[width=.45\linewidth]{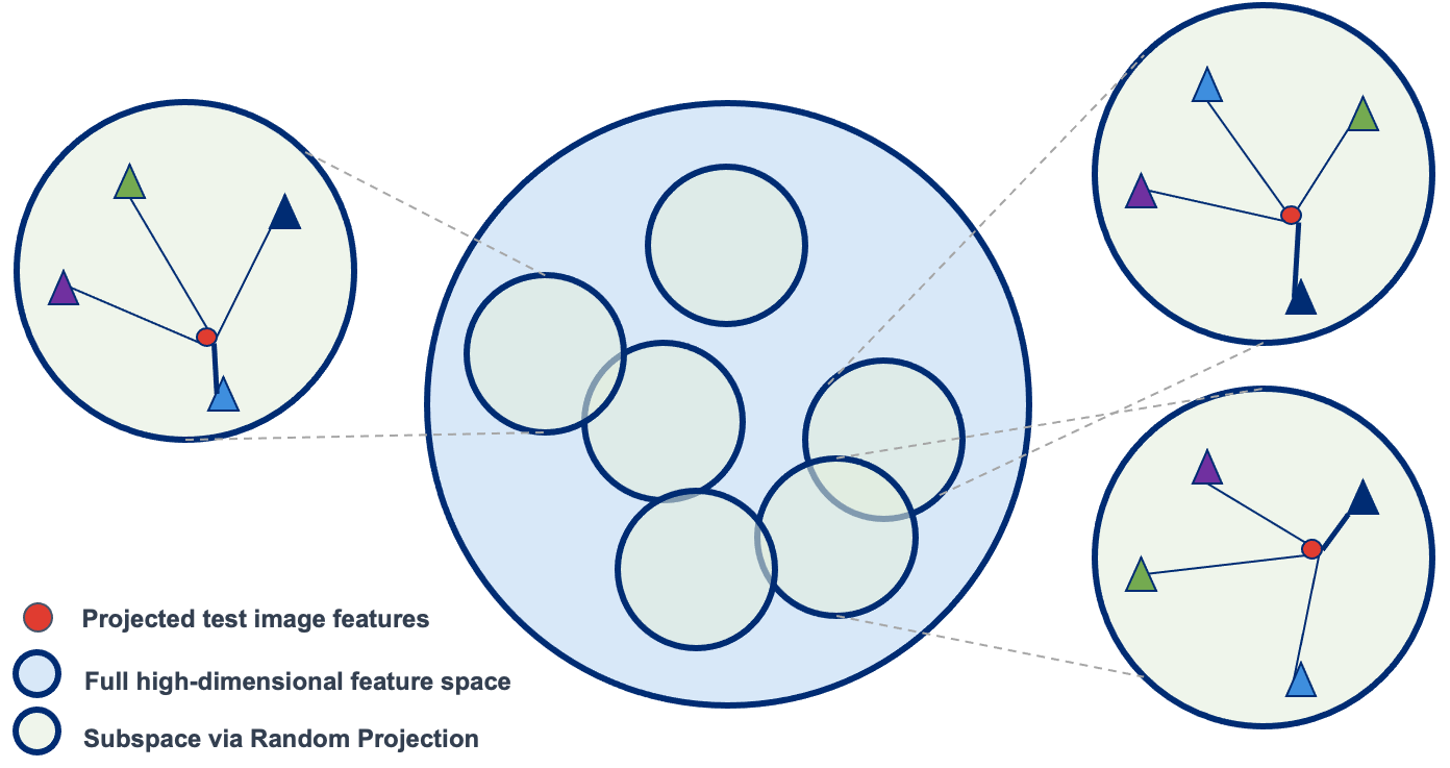} 
    \end{tabular}
\end{center}
  \caption{\textit{Adversarial example detection via Random Subspace Analysis.} Random projections provide a mechanism for studying activation behavior with the aim to detect clean vs. adversarial examples. The likely existence of \textit{non-robust but useful features} is exploited by measuring the consistency of the activation and latent space properties across multiple subspaces. Shown above, the two subspaces on the right are consistent with respect to which class prototype (colored triangles) is closest to the test point (red circle). The leftmost subspace is inconsistent since a different colored class prototype is preferred.}
\label{fig:rsa_graphic}
\vspace{-4mm}
\end{figure*}

\section{Related Work}

While developing novel attack and defense strategies occupies a large proportion of current and past AML research work, interest in automated detection of adversarial examples has grown considerably in recent years.  Adversarial detection is a complementary problem to adversarial defense, but we focus here on efforts to explicitly predict cases of adversarial manipulation and leave it to downstream methods to mitigate the effects of any such manipulations.

We find the bulk of the AE detection work falls roughly into the following categories:

\begin{itemize}
    \itemsep0em
    \item \textbf{Input-space detection} - Methods that seek to identify adversarial examples through an examination/manipulation of the inputs to DNNs \cite{liang2018detecting, tian2018detecting, xu2017feature}.
    \item \textbf{Feature-space detection} - Methods which typically train new detector models on intermediate or final DNN layer outputs \cite{metzen2017detecting, bagnall2017training, crecchi2019detecting, wang2019evaluating, fidel2019explainability, abdelzaher2018toward, miller2017not, papernot2018deep, roth2019odds, ma2018characterizing, feinman2017detecting}.
    \item \textbf{Robustness Certificates} - Theoretical proofs for DNN robustness under specific threat models (assumptions for both attacker and defender) \cite{wong2018provable, raghunathan2018certified, weng2018towards, weng2019proven, wang2019evaluating, singla2019robustness, levine2019wasserstein, lyu2019fastened, kumar2020curse, lecuyer2019certified, boopathy2019cnn, gehr2018ai2}.
    \item \textbf{Statistical Detection} - Methods that focus on generating robust test statistics that separate clean from adversarial examples \cite{li2017adversarial, grosse2017statistical, quintanilha2018detecting}.
    \item \textbf{Network Coverage} - Methods that examine neuron coverage as a means to test DNN behavior and potentially establish differences between clean and adversarial examples \cite{ma2018deepgauge, ma2018combinatorial, sun2018concolic, sun2018testing, pei2017deepxplore}.
\end{itemize}

Of specific relevance and similarity to our approach, are a set of high-performing methods~\cite{lee2018simple, ma2018characterizing, feinman2017detecting, papernot2018deep} which use local latent space geometry for AE detection. The work in~\cite{feinman2017detecting} originally combined kernel density estimation and Bayesian uncertainty estimation to identify AEs as predicted-class outliers. Similarly,~\cite{lee2018simple} examined detection through the lens of Gaussian Discriminant Analysis to estimate class confidence scores based on Mahalanobis distance (combined with a layer-wise logistic regression model to predict AE likelihood). 

Both~\cite{ma2018characterizing, papernot2018deep} perform AE detection using nearest-neighbor principles where~\cite{ma2018characterizing} relies on Local Intrinsic Dimensionality (LID) to characterize local neighborhood structure while~\cite{papernot2018deep} uses discrete k-Nearest Neighbors along with conformal prediction to identify neighborhood (in)consistencies.  Our approach is inspired by elements of these papers and in particular, the approach towards AE detection by identifying (in)consistencies in latent-space behavior for both adversarial and clean examples.

\section{Novel contributions}
We present a novel approach to AE detection which leverages random projection subspace analysis to differentiate between model behavior under clean and adversarial conditions. Additional advantages and novel features of our proposed approach are as follows:

\begin{itemize}
    \itemsep0em
    \item Our detection scheme uses random projection subspaces to compare and analyze network activation behavior in a self-supervised manner.
    \item While most state of the art (SOTA) methods require some prior knowledge of attacks to train their detectors, our method is truly agnostic to attack strategy and goal, and requires a calibration set composed solely of clean examples.
    \item Additionally, while most detection methods rely on large calibration datasets, our method is extremely data efficient and requires only a fraction of the calibration data typically used. Our experiments show that calibration data requirements scale according to the number of classes rather than the method itself.
    \item Last, we evaluate our method under rigorous constraints by significantly reducing the size of the calibration set and restricting its composition to clean examples only (a true attack-agnostic paradigm). Under these more realistic and difficult conditions, our method far outperforms comparable methods across a variety of unseen attacks.
\end{itemize}

\section{Background and Notation}
We first provide some basic notation and assumptions.  In typical image classification tasks, we start from a set of exemplars $(x,y)$ where $x \in \mathbb{R}^{C\times W \times H}$ and $y\in Y=\{1,...,K\}$ (i.e., $K$ classes).  The aim is to produce a model, $f:X\longrightarrow Y$, which minimizes a loss, $L(f(x;\theta),y)$.  The model is typically composed of a set of computational layers parameterized in aggregate by weight parameters $\theta$ giving us $\hat{y}=f(x;\theta)$.  Assuming a purely feedforward model (a common form for image-based DNNs), we can decompose $f$ as follows:

\vspace{-0.5cm}
\begin{equation}
    \begin{split}
        &f(x;\theta) = f_L^{\theta_L}\circ f_{L-1}^{\theta_{L-1}}\circ \dots f_2^{\theta_2}\circ f_1^{\theta_1}(x) = \hat{y} \\
        &\text{where}~f_l^{\theta_l}(\cdot) = z_l
    \end{split}
\end{equation}

Note that for machine vision problems, $f(\cdot)$ is typically implemented as a deep neural network, whereby the intermediate layer activations, $z_l$ may be 2D vectors or 3D tensors. Also, note that typically $f(\cdot)$ produces an output that represents a distribution over the $N$ classes where $\hat{y}=\argmax z_L$. During a forward pass of our model, we record and aggregate all intermediate activations into $\pmb{z}=\{z_1,\dots,z_L\}$.

\section{Threat Model}
We define an adversarial example as a perturbation, $\eta$, such that for a correctly classified clean input, $f(x;\theta)=\hat{y}=y_{true}$, whereas the added perturbation yields an incorrect classification, $f(x+\eta;\theta)=\hat{y}\neq y_{true}$.  In targeted attacks cases, the AE is constructed to cause $\hat{y}=y_{target}$ whereas in  untargeted scenarios, the AE is constructed to cause $\hat{y}\neq y_{true}$. In either case, the representation of the attacked image in feature space shifts away from its clean counterpart as illustrated in Figure~\ref{fig:feature_space}.

In order to construct the most challenging detection task as possible from the perspective of the defender, we consider white-box scenarios whereby the attackers have full knowledge and access to the network weights and architecture to be attacked. In this setting, we assume that the attacker has access to the clean input and may modify it within specified constraints (e.g., a bound on the $L_p$-norm, i.e., $||\eta||_p \leq \epsilon$).  

\section{Adversarial Attack Detection Method}

\subsection{Dimensionality Reduction via Random Projections}
\label{sub:rp}
Recent work by~\cite{ilyas2019adversarial} provides evidence for the existence of \textit{robust} and \textit{non-robust but useful features} which contribute to task performance but crucially are informative regarding the susceptibility of the DNN to attacks. One possible mitigation strategy for adversarial manipulation of non-robust features is to perform dimensionality reduction of learned representations whereby the resulting features (from the \textit{robust} set) preserve classification-relevant information and reduce or remove the impact of manipulated features (from the \textit{non-robust} set). 

Conventional dimensionality reduction techniques (e.g., Principal Components Analysis, Singular Value Decomposition, Independent Component Analysis) offer principled approaches to achieving the reduction, but a primary disadvantage is that they are computationally expensive to compute in high-dimensions. Furthermore, while beneficial in many scenarios, these methods produce a single deterministic subspace which limits the scope for comparative analysis between properties of the reduced and full representations. 

Instead, Random Projections (RP) do not have the aforementioned drawbacks, provide many desirable properties (explained later), and allow for more flexible analysis.  Our approach will be to project layer activations into a series of random subspaces and then compare the relation of these reduced dimension projections with respect to class prototypes (in each subspace). Differences between clean and adversarial behavior will be exploited to produce the final detection. Definitions of these terms and details will be discussed in the following sections. 

First, define a random projection of activation $z_i$ as:
\begin{equation}
    \hat{z}_i = Rz_i
\end{equation} 
where $R \in \mathbb{R}^{k \times d}$ ($k \ll d$), $z_i \in \mathbb{R}^{d}$, and $\hat{z}_i \in \mathbb{R}^{k}$, each $R_{ij}$ is sampled independently from a standard normal distribution, and where representations are extracted from layer $i \in \{l~|~1 \leq l \leq L\}$ of the DNN. Strictly speaking, the rows of $R$ should be orthogonal. However, if $R$'s elements are sampled independently, it can be shown that the rows are approximately orthogonal (\cite{bingham2001random, hecht1994context}) and consequently the projections are reasonably approximate as well. 

\subsection{Theoretical Considerations}
\label{sub:jl}

We draw particular motivation for using RP from the Johnson-Lindenstrauss Lemma which guarantees that with high probability, distances between pairs of points in the random subspace are preserved up to a scaling $(1\pm\epsilon)$. 

\vspace{1mm}
\begin{lemma}[Johnson-Lindenstrauss~\cite{johnson1984extensions}]
\hfill \break Let $X = \{x_1,...,x_k\}$ in $\mathbb{R}^n$.  There exists a random function $f:\mathbb{R}^n \longrightarrow \mathbb{R}^m$ such that for any pair of points $x_i,x_j$

\begin{equation}
    (1-\epsilon)||x_i-x_j||^2_2 \leq ||f(x_i) - f(x_j) ||^2_2 \leq (1+\epsilon)||x_i-x_j||^2_2
\end{equation}
with probability at least $\frac{1}{k}$ so long as $m \geq \frac{8\log{k}}{\epsilon^2}$.

\label{lem:jl_lemma}
\end{lemma}

A proof of Lemma~\ref{lem:jl_lemma} can be found in~\cite{dasgupta1999elementary}.  

The J-L Lemma is a statement about the existence of (random) embeddings in lower dimensions whereby the distortion of distances between a pair of points can be guaranteed to be bounded. This forms the basis for our detection method as it allows comparison between the proximity of points in the ambient feature space and in arbitrary random subspaces. Section~\ref{sub:rsa} will describe how we can use random subspaces to check whether distances between suspected adversarial examples and known class prototypes remain consistent across random subspaces (with inconsistencies signalling a potential AE). The J-L Lemma guarantees that these distances should be preserved. 

It is important to note that while the bound is tight, it represents a worst case scenario.  In practice, distances are well-preserved below the theoretical value of $m$ (see~\cite{bingham2001random, dasgupta2013experiments}). Prior work has also reported that RP can help preserve other aspects of geometry such as cluster separability~\cite{dasgupta2013experiments}, volume, and distances to affine spaces~\cite{magen2002dimensionality}, both theoretically and empirically.  Connections between Restricted Isometry Property (RIP), a key ingredient for compressive sensing, and the J-L lemma were also made in~\cite{krahmer2011new}.

\subsection{Random Subspace Analysis}
\label{sub:rsa}
Our approach focuses on measuring the consistency of the local geometry around the test image embedding between the ambient and subspace representations. In particular, we leverage the ability to produce arbitrary random projections to compare DNN activations in multiple subspaces derived from the original full-dimensional space (illustrated conceptually in Figure~\ref{fig:rsa_graphic}). 

More formally, define a set of $M$ projections, $\{R_m | 1 \leq m \leq M\}$ applied to activation $z_l$ from layer $l$ to produce a set of subspace representations:

\begin{equation}
    \pmb{\hat{z}_l} = \{\hat{z}_{l,m}\} = \{R_m z_l ~|~ 1 \leq m \leq M\}
\end{equation}

\noindent where elements of $R_m$ are sampled independently from a standard normal distribution. When the random projection subspace is large, the projection vectors will tend to become normal to each other as the dimension increases.  As shown in~\cite{bingham2001random, hecht1994context}, any sampling of $R$ will be nearly orthonormal, ensuring that J-L Lemma holds in practice for arbitrary sampling using the standard normal distribution. 

Because the RP preserves distances within some scaling, the potential loss of information can be advantageous.  As evidenced by nearest-neighbor type analyses (e.g.,~\cite{papernot2018deep}), adversarial features are likely to exhibit greater similarity to exemplars of the non-true class than the true class.  

The existence of \emph{robust} and \emph{non-robust but useful features}, suggests that the non-robust features may be more likely than others to be manipulated under attack. Under a set of random projections, the influence of the manipulated features on the final outcome becomes increased/decreased as a function of the specific projection itself. 

In the case that the manipulated features become less influential in the random subspace, the projection should show greater similarity to examples of the true class than the incorrect classes.  Conversely, if the manipulated features become more influential, then the projection should show greater dissimilarity with the true class representation. When examined over a set of random projections, inconsistent behavior is more likely to be exposed and amplified. For clean examples, projections should exhibit strong overall self-consistency to the true class across a set of projections (as suggested by the J-L lemma).

To build our detection model, we first extract layer-wise features $z_l$ for each image $x$ in a training dataset of clean images $\mathcal{D_T} = \{(x,y) | x \in \mathbb{R}^{C \times W \times H}\}, y \in \{1,...,K\}$. We compute class-conditional prototypes: 
\begin{equation}
    \mu_{l,k} = \frac{1}{N_k} \sum_{i:~y_i=k}{f_l(x_i;\theta)}    
\end{equation}
where $N_k$ is the number of examples with class label $k$.  Then, as described in Algorithm~\ref{alg:rp_analysis}, we project both the test point features and the prototypes into the same subspace, determine the nearest prototype in the subspace, and return the associated label for each layer of interest.

\begin{algorithm}[h!]
    \small
    \KwResult{Random Subspace Analysis}
    \SetKwInOut{Input}{Input}
    \SetKwInOut{Output}{Output}
    \Input{
        $z_l$ : Activation for the test image at layer $l$ \\
        $\pmb{\mu}_{l}$ : Class prototypes (k) at layer $l$ \\
        $\pmb{R}_l$ : Set of $M$ random projection matrices at layer $l$ \\
        $d(\cdot)$ : Distance function (i.e., euclidean, cosine)
    }
    \Output{
        $\mathcal{P}_l$ : Set of $M$ predicted class labels (one per RP)
    }
    
    Initialize the RP nearest-label array $\mathcal{P}_l = [~]$\;
    \For{$m \in \{1, ..., M\}$}{
        Initialize distances array $D = [~]$\;
        Project activation: $\hat{z}_l = R_m z_l$\;
        \For{$k \in \{1,...,K\}$}{
            Project prototype: $\hat{\mu}_{l,k} = R_m \mu_{l,k}$\;
            Add $d(\hat{z}_l, \hat{\mu}_{l,k})$ to $D$\;
        }
        Add $c = \argmin_k(D)$ to $\mathcal{P}_l$\;
    }
    \Return $\mathcal{P}_l$\;
    \caption{Random Subspace Analysis}
    \label{alg:rp_analysis}
\end{algorithm}
\vspace{-4mm}

\subsection{AE Detection}
We next apply a simple decision rule on the nearest-prototype set $\mathcal{P}_l$ (which contains the label of the nearest prototype to the test point in each subspace), to predict whether the test point is adversarial (represented as $\hat{a}$) and where $mode(\mathcal{P}_l)$ represents the label occurring most frequently in $\mathcal{P}_l$.

\begin{equation}
\hat{a} = 
    \begin{cases}
  1 & \text{if $\frac{1}{M} \sum\limits_{c\in\mathcal{P}_l}\mathbbm{1}(c=mode(\mathcal{P}_l)) < \alpha$ for $\alpha \in [0, 1]$} \\ 
  0 & \text{otherwise}
\end{cases}
\label{eq:dec_rule}
\end{equation}
Note that the set $\mathcal{P}_l$ may be replaced by $\mathcal{P}$ which is simply an aggregation of label preference across multiple layers of the network (i.e, $\mathcal{P} = \bigcup \mathcal{P}_l$ for $l \in L'\subseteq L$). Essentially, Eq.~\ref{eq:dec_rule} represents the fraction of label agreement across all random subspaces (as determined by the distance from the test point to its nearest prototype).

Of course, more powerful decision rules may be developed, but the rule in Eq. (\ref{eq:dec_rule}) provides several notable advantages: (1) it is interpretable and simple to compute, (2) it is tunable by a single hyperparameter $\alpha$, (3) it is agnostic to the ground-truth label, and (4) it is non-differentiable (increasing its robustness against many powerful attacks).  Note that $\alpha=0$ corresponds to the degenerate case where $x$ is always labeled as clean. In contrast, $\alpha=1$ is the most stringent case such that the label for the nearest-prototype must be the same for all random projections in order for $x$ to be considered clean.  To be truly attack agnostic, the selection of $\alpha$ can be performed absent of any attack data (e.g., to reduce false alarms).  However, if attack data is available, $\alpha$ may be tuned in a more principled way.  

The benefit from (3) above is of particular importance given that the true class is not known for new test points, and thus we cannot condition the detection on the true label at inference. While many methods condition the detection on the DNN's prediction $\hat{y}$ (e.g.,~\cite{papernot2018deep, ma2018characterizing, feinman2017detecting}), our approach considers only whether the label of the nearest prototype stays consistent across random subspaces. Thus, we are able to use class-level representations while still separating AE detection from image classification. In essence, our method takes advantage of competing elements of adversarial attacks, namely changing the predicted label of the model without changing the true meaning of the input.

\begin{figure}[t!]
\begin{center}
    \begin{tabular}{c}
        \includegraphics[width=\linewidth]{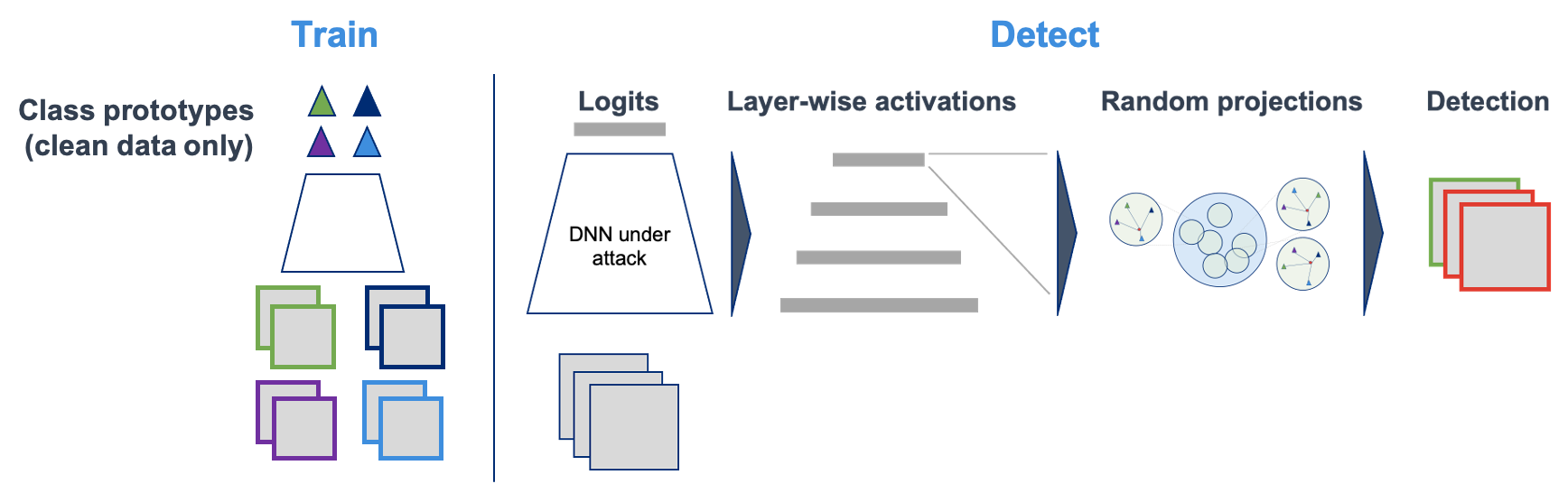}
    \end{tabular}
\end{center}
  \caption{\textit{AE detection pipeline:  Training: class prototypes are computed from clean data only (one prototype per class).  Detection: activations computed for the test images (gray squares) are projected to a set of random subspaces and consistency is measured in terms of label preference.}}
\label{fig:rsa_pipeline}
\vspace{-4mm}
\end{figure}

\section{Data and Models}
\subsection{Adversarial Example Generation}
For evaluation purposes only, we generated sets of ({\it clean, adversarial}) image pairs on which to evaluate. We make use of the Foolbox toolkit~\cite{rauber2017foolbox} to generate attacks with varying complexity and strategy, including FGSM~\cite{goodfellow2014explaining}, PGD~\cite{madry2017towards}, CWL2~\cite{carlini2017towards}, JSMA~\cite{papernot2016limitations}, EAD~\cite{chen2017ead}, and gradient-free noise).  We use the respective published attack hyperparameters during attack generation. 

We evaluate our detection approach 
(Figure~\ref{fig:rsa_pipeline}) 
under a white-box threat model whereby the adversary has full knowledge of the network architecture and weights. Attack parameters are chosen to be consistent with those commonly found in the literature (which balance attack effectiveness and subtlety). This threat model provides an approximate lower-bound for detection given that the adversary should not be able to improve their attack under less-favorable conditions.

\subsection{Data}
For each experiment, we generate a set of 4000 ({\it clean, adversarial}) image pairs for each attack under consideration for CIFAR10, SVHN, and mini-ImageNet.  To fit our detection model, we reserve 10\% of the clean examples only for computing the class-specific prototypes and the remaining 90\% of exemplars for evaluation.  

For each attack, we follow standard practice for evaluating AE detection and evaluate only the ({\it clean, adversarial}) pairs where both elements achieve their objectives (i.e., correct classification, misclassification). In both cases, we avoid confounding our detector's performance with other data/model-related issues that would come from misclassification of clean examples or failed attempts to attack the model.

\subsection{White-box Models}
We focus our efforts on attacking commonly used, high-performing architectures for image classification.  We train ResNet[18,34]~(\cite{he2016deep}) and DenseNet161~(\cite{huang2017densely}) models from scratch on each dataset. We train for 200 epochs using SGD with a base learn rate of $10^{-1}$ (decreasing by 0.1 every 50 epochs starting from epoch 100), momentum of 0.9, and weight decay constant of $10^{-4}$.  These models are then used directly for the white-box attack generation to ensure the most challenging AE detection scenario.

\subsection{Baselines and SOTA Comparisons}
We evaluate our method against SOTA approaches from~\cite{lee2018simple} and~\cite{ma2018characterizing} which use local latent space geometry for adversarial detection. These methods are not attack agnostic, so we must modify these approaches to fit our paradigm. First, we are primarily interested in attack-agnostic detection so we must have detection methods that do not require any attack data during the detector calibration step. Second, we focus on a test paradigm where the calibration set is limited in size which emphasizes advantages of the method relative to any performance gained from more data.  Modifying these baselines further allows better insight into whether the performance of those methods is attributed to the method itself or the access to ground truth attack data during training.

{\bf DMD-OC:} The Deep Mahalanobis Detector (DMD)~\cite{lee2018simple} applies Gaussian Discriminant Analysis to compute class-conditional confidence scores across a set of DNN layers. The maximum score over all classes is taken at each layer and then a weighted sum of the scores is used to produce the final output. The original method trained a logistic regression classifier on clean and adversarial data using the confidence scores as features to learn an optimal weighting of layer-specific scores.  

{\bf LID-OC:} The Local Intrinsic Dimensionality (LID)~\cite{ma2018characterizing} detector estimates the local intrinsic dimensionality of the test sample from its k-nearest neighbors (we use $k=20$ as in the original paper) based on activations at selected layers in the DNN.  Similar to DMD, the original method trained a logistic regression classifier on clean and AE data using the LID estimates across layers as features to predict if the test point is clean or adversarial. 

For these baselines, in order to fit the attack-agnostic paradigm where adversarial data for training is not available, we replace the Logistic Regression classifiers in both methods with a One-Class Support Vector Machine (SVM)~\cite{scholkopf2001estimating} trained only on clean activations. We do not modify the computation of the confidence scores or LID values, and we use the same DNN layers for all baseline detection methods.

\section{Experiments}
In all of the following experiments, we run our detection method using the output from the last layer prior to the classification layer.  Following standard practice, the primary performance metric is ROC Area Under the Curve (AUC) score which allows us to sweep over decision thresholds (i.e., $\alpha$ for our method and similar for other baselines).

\subsection{White-Box Attacks}
We first consider AE detection for untargeted white-box attacks.  Since our method is agnostic to the attack algorithm, we observe the detection performance across a range of attacks and consider the differences in performance between them. Table~\ref{tab:f1_whitebox_untargeted} illustrates the detection performance for untargeted attacks.  In all cases, $k=16$ for the RP dimension, $M=8$ projections were used for the detector.

\begin{table*}[t!]
\centering
\caption{AUC score for detection of white-box untargeted perturbation attacks on CIFAR10 and SVHN. Best methods indicated in \textbf{bold}.}
\label{tab:f1_whitebox_untargeted}

\resizebox{12cm}{!}{
\begin{tabular}{l|l|l|c|c|c|c|c|c}
     &             &  & \multicolumn{6}{c}{Attack} \\ 
Dataset & Architecture & Detector & FGSM &  PGD &  JSMA &  EAD & CWL2 &  Noise \\
\hline
\multirow{9}{*}{CIFAR10} & \multirow{3}{*}{ResNet18} & LID-OC &   0.570 &  0.525 &   0.547 &  0.509 &   0.540 &    0.533 \\
     &             & DeepMD-OC &   0.502 &  0.550 &   0.547 &  0.547 &   0.551 &    0.553 \\
     &             & DeepRP (ours) &   \textbf{0.955} & \textbf{0.964} &   \textbf{0.962} &  \textbf{0.951} &   \textbf{0.954} &    \textbf{0.930} \\
\cline{2-9}
     & \multirow{3}{*}{ResNet34} & LID-OC &   0.495 &  0.522 &   0.519 &  0.512 &   0.498 &    0.505 \\
     &             & DeepMD-OC &   0.500 &  0.601 &   0.594 &  0.585 &   0.583 &    0.602 \\
     &             & DeepRP (ours) &   \textbf{0.946} &  \textbf{0.974} &   \textbf{0.972} &  \textbf{0.941} &   \textbf{0.966} &    \textbf{0.915} \\
\cline{2-9}
     & \multirow{3}{*}{DenseNet161} & LID-OC &   0.491 &  0.496 &   0.533 &  0.493 &   0.486 &    0.553 \\
     &             & DeepMD-OC &   0.500 &  0.500 &   0.500 &  0.500 &   0.500 &    0.500 \\
     &             & DeepRP (ours) &   \textbf{0.928} &  \textbf{0.926} &   \textbf{0.931} &  \textbf{0.920} &   \textbf{0.926} &    \textbf{0.952} \\
\cline{1-9}
\cline{2-9}
\multirow{9}{*}{SVHN} & \multirow{3}{*}{ResNet18} & LID-OC &   0.601 &  0.482 &   0.480 &  0.487 &   0.468 &    0.299 \\
     &             & DeepMD-OC &   0.501 &  0.603 &   0.569 &  0.603 &   0.584 &    0.571 \\
     &             & DeepRP (ours) &   \textbf{0.954} &  \textbf{0.957} &   \textbf{0.965} &  \textbf{0.951} &   \textbf{0.952} &    \textbf{0.950} \\
\cline{2-9}
     & \multirow{3}{*}{ResNet34} & LID-OC &   0.543 &  0.509 &   0.514 &  0.494 &   0.475 &    0.384 \\
     &             & DeepMD-OC &   0.507 &  0.591 &   0.580 &  0.595 &   0.592 &    0.577 \\
     &             & DeepRP (ours) &   \textbf{0.953} &  \textbf{0.937} &   \textbf{0.956} &  \textbf{0.944} &   \textbf{0.944} &    \textbf{0.923} \\
\cline{2-9}
     & \multirow{3}{*}{DenseNet161} & LID-OC &   0.513 &  0.581 &   0.585 &  0.588 &   0.569 &    0.795 \\
     &             & DeepMD-OC &   0.500 &  0.500 &   0.500 &  0.500 &   0.500 &    0.500 \\
     &             & DeepRP (ours) &   \textbf{0.959} &  \textbf{0.972} &   \textbf{0.975} &  \textbf{0.967} &   \textbf{0.967} &    \textbf{0.968} \\
\hline
\end{tabular}
}
\vspace{-2mm}
\end{table*}

\subsection{Sensitivity Studies}
We run additional sensitivity studies to determine the impact of both the dimensionality of the random subspace as well as the number of subspaces considered during the detection process.  

{\bf RP Dimension Dependency}
Table~\ref{tab:k_dependency_18} shows the impact of changing the dimension of the RP on the AE detection result.  In all cases, $M=8$ projections were used for the layer prior to the final output layer. The full dimensionality of the final layer is 512, so the subspace dimensions studied here represent a modest-to-aggressive reduction.

{\bf Dependence on Number of Projections}
Table~\ref{tab:M_dependency_18} captures the impact of the number of projections, $M$, considered at detection time.  We fix the dimensionality to $k=16$.  We focus here on a range of projections to capture the tradeoff between having sufficient data for measuring label consistency and over-inflating the computational requirements for performing detection.

{\bf Expanded Calibration Data}
We perform additional experiments which increase the amount of clean-only calibration data available to the detectors and present them in Table~\ref{tab:extended_training}.  Since the detectors are only fit to clean data and due to the limited size of the calibration/test splits for detection, we expand the available data by drawing from the original training set used for training the core model.  While this is not common practice, this approach still has practical value since it allows the detectors to use prototypes based on embeddings the network was directly trained to produce.  Data used for detection is held out as normal. 

These results suggest that the strength of other detection methods scales in proportion to the available calibration data whereas our method provides consistent performance for small and larger training set sizes.

{\bf Targeted Attacks}
Table~\ref{tab:targeted} captures the performance of the proposed method on targeted attacks.  A target label was chosen at random from the CIFAR10 set and used to generate targeted attacks. The detector was calibrated on 20\% of the clean samples. Results indicate that the proposed method significantly outperforms baselines even when the attack is steered towards a particular label.

\subsection{Experiments on mini-ImageNet}
We run additional experiments on mini-ImageNet to test the effectiveness of our method on more challenging data (Table~\ref{tab:mini_imagenet}).  We focus on scaling the complexity of the detection task to the image data (higher resolution, greater diversity) rather than the classification task itself (e.g., more classes).  For these experiments, we choose the DeepRP hyperparameters to be $M=32$ and $k=16$. We also expand the calibration dataset to be 2000 clean examples.

Results show that our method far outperforms all competing methods even as the images become more complex in terms of resolution and statistics.  That said, we believe that the difficulty of the ImageNet data relative to CIFAR or SVHN motivates a greater expansion of the calibration dataset to get more stable estimates of the class prototypes.

\section{Discussion}
Experiments demonstrate the efficacy of our approach and leads to several key observations.  First, the overall detection performance of our method is consistently high across attack algorithms, demonstrating a degree of independence between our detection method and the details of the attack (under the same threat model).  Slight differences occur between the detection performance given the architecture under attack, but the detection performance is consistent.

Our method also outperforms the other baselines given limited access to clean-only training data.  This performance difference is mostly attributed to the significant reduction in the calibration set size.  While all methods benefit greatly from having more calibration data (i.e., to produce more stable prototypes (ours), reduce the sparsity of the latent space (LID), improve the mean/covariance estimates (DMD)), results indicate that our approach is considerably more robust under this constraint. This lends credence to the use of our approach under a greater range of conditions where the attack is unknown and available calibration data may be limited.

Tables~\ref{tab:k_dependency_18} and~\ref{tab:M_dependency_18} indicate some dependence on both the dimension of the random projection ($k$) and the number of projections considered during the detection process ($M$). There is a clear tradeoff between choices of $k$ and $M$. For $k$, results suggest that the RP must perform a certain degree of representation compression, but too little compression results in a representation too similar to the original (i.e., inclusion of too many non-robust features) and too much results in a significant loss of information (i.e., loss of robust \emph{and} non-robust features).  For $M$, the balance must be between too few subspaces under consideration (i.e., resulting in insufficient observations of activation behavior) and too many (i.e., where subspace diversity becomes less prominent).

Furthermore, the additive noise attack produces, in general, the lowest detection performance. Since this attack requires no knowledge of the underlying model architecture or weights, it is unable to preferentially target (implicitly or explicitly) the non-robust features over the robust ones. As such, the random subspaces are more likely to produce consistent behavior and thereby miss possible detections. While this attack is notably weak compared to more sophisticated approaches (e.g.,\cite{carlini2017towards}), it highlights how an adaptive adversary might attempt to circumvent our detection scheme. However, while modifying an attack to more evenly manipulate features may evade detection, it may have unintended and detrimental consequences on the overall viability of the attack itself. We leave it to future work to investigate this tradespace of competing objectives in more depth.

Lastly, recent literature~\cite{bulusu2020anomalous, hsu2020generalized, fort2021exploring, ndiour2020out} has provided new methods for detecting out-of-distribution or anomalous samples wherein adversarial examples may be viewed as a special case.  Specifically, recent and concurrently developed self-supervised~\cite{mohseni2020self} and unsupervised~\cite{ko2021unsupervised, koner2021oodformer} methods in those problem domains present a promising direction for future research.

\begin{table*}[t!]
\centering
\caption{AUC score for detection of white-box untargeted perturbation attacks on CIFAR10, SVHN as the dimension of the RP changes. Best results indicated in \textbf{bold face}.}
\label{tab:k_dependency_18}
\resizebox{11cm}{!}{ 
\begin{tabular}{l|l|l|c|c|c|c|c|c}
     &             &  & \multicolumn{6}{c}{Attack} \\   
Dataset & Architecture & Detector & FGSM &  PGD &  JSMA &  EAD & CWL2 &  Noise \\
\hline
\multirow{5}{*}{CIFAR10} & \multirow{5}{*}{ResNet18} & DeepRP-k8 &   0.935 &  0.948 &   0.947 &  0.940 &   0.928 &    \textbf{0.948} \\
     &          & DeepRP-k16 &   0.955 &  \textbf{0.964} &   \textbf{0.962} &  \textbf{0.951} &   \textbf{0.954} &    0.930 \\
     &          & DeepRP-k32 &   \textbf{0.961} &  0.959 &   0.961 &  0.943 &   0.945 &    0.901 \\
     &          & DeepRP-k64 &   0.952 &  0.958 &   0.945 &  0.926 &   0.931 &    0.843 \\
     &          & DeepRP-k128 &   0.910 &  0.918 &   0.904 &  0.855 &   0.873 &    0.752 \\
\cline{1-9}
\cline{2-9}
\multirow{6}{*}{SVHN} & \multirow{5}{*}{ResNet18} & DeepRP-k8 &   0.953 &  \textbf{0.958} &   \textbf{0.966} &  \textbf{0.961} &   \textbf{0.959} &    \textbf{0.967} \\
     &          & DeepRP-k16 &   \textbf{0.954} &  0.957 &   0.965 &  0.951 &   0.952 &    0.950 \\
     &          & DeepRP-k32 &   0.934 &  0.916 &   0.943 &  0.909 &   0.922 &    0.904 \\
     &          & DeepRP-k64 &   0.877 &  0.845 &   0.879 &  0.838 &   0.848 &    0.834 \\
     &          & DeepRP-k128 &   0.824 &  0.779 &   0.811 &  0.775 &   0.783 &    0.768 \\
\hline
\end{tabular}
}
\end{table*}

\begin{table*}[t!]
\centering
\caption{AUC score for detection of white-box untargeted perturbation attacks on CIFAR10, SVHN as the number of projections changes. Best results indicated in \textbf{bold face}.}
\label{tab:M_dependency_18}
\resizebox{11cm}{!}{ 
\begin{tabular}{l|l|l|c|c|c|c|c|c}
     &             &  & \multicolumn{6}{c}{Attack} \\ 
Dataset & Architecture & Detector & FGSM &  PGD &  JSMA & EAD &  CWL2 &  Noise \\
\hline
\multirow{5}{*}{CIFAR10} & \multirow{5}{*}{ResNet18} & DeepRP-M2 &   0.768 &  0.770 &   0.771 &  0.750 &   0.764 &    0.786 \\
     &          & DeepRP-M4 &   0.895 &  0.911 &   0.910 &  0.889 &   0.904 &    0.891 \\
     &          & DeepRP-M8 &   0.955 &  0.964 &   0.962 &  0.951 &   0.954 &    0.930 \\
     &          & DeepRP-M16 &   0.968 &  0.976 &   0.974 &  0.966 &   0.965 &    0.953 \\
     &          & DeepRP-M32 &   \textbf{0.974} &  \textbf{0.980} &   \textbf{0.980} &  \textbf{0.971} &   \textbf{0.970} &    \textbf{0.966} \\
\cline{1-9}
\cline{2-9}
\multirow{5}{*}{SVHN} & \multirow{5}{*}{ResNet18} & DeepRP-M2 &   0.767 &  0.753 &   0.770 &  0.761 &   0.767 &    0.823 \\
     &          & DeepRP-M4 &   0.904 &  0.901 &   0.920 &  0.894 &   0.895 &    0.910 \\
     &          & DeepRP-M8 &   0.954 &  0.957 &   0.965 &  0.951 &   0.952 &    0.950 \\
     &          & DeepRP-M16 &   0.966 &  0.972 &   0.974 &  0.964 &   0.967 &    0.960 \\
     &          & DeepRP-M32 &   \textbf{0.970} &  \textbf{0.977} &   \textbf{0.979} &  \textbf{0.972} &   \textbf{0.976} &    \textbf{0.971} \\
\hline
\end{tabular}
}
\end{table*}

\section{Conclusions}
We present a novel approach to adversarial example detection via random subspace analysis. We use random projections to reduce dimensionality of deep features and then examine the consistency of features across a set of subspaces to detect attacks. We evaluate our method under much more rigorous constraints than prior approaches by constraining the calibration set to clean examples only. 

Our results demonstrate that our approach, while being agnostic to the attack strategy or objective (i.e., targeted/untargeted), consistently outperforms against a range of SOTA attack strategies with various degrees of sophistication. We believe that this work opens up new approaches for analyzing deep features in the context of adversarial example detection via random subspace analysis.

\begin{table*}[t!]
\centering
\caption{AUC score for detection of white-box untargeted perturbation attacks on CIFAR10 for 10x more training data. Best methods indicated in \textbf{bold face}.}
\label{tab:extended_training}
\resizebox{11cm}{!}{ 
\begin{tabular}{l|l|l|c|c|c|c|c}
     &             &  & \multicolumn{5}{c}{Attack} \\   
Dataset & Architecture & Detector & FGSM &  PGD &  JSMA &  CWL2 &  Noise \\
\hline
\multirow{6}{*}{CIFAR10} & \multirow{3}{*}{ResNet18} & LID-OC &   0.495 &  0.525 &   0.525 &   0.535 &    0.763 \\
        &          & DeepMD-OC &   0.833 &  0.864 &   0.885 &   0.860 &    0.849 \\
        &          & DeepRP (ours) &   \textbf{0.961} &  \textbf{0.966} &   \textbf{0.963} &   \textbf{0.957} &    \textbf{0.948} \\
\cline{2-8}
        & \multirow{3}{*}{ResNet34} & LID-OC &   0.486 &  0.493 &   0.584 &   0.556 &    0.739 \\
        &          & DeepMD-OC &   0.837 &  0.899 &   0.904 & 0.902 &    0.861 \\
        &          & DeepRP (ours) &   \textbf{0.965} &  \textbf{0.973} &   \textbf{0.971} &   \textbf{0.967} &    \textbf{0.928} \\
\hline
\end{tabular}
}
\end{table*}

\begin{table*}[t!]
\centering
\caption{AUC score for detection of white-box untargeted perturbation attacks on mini-Imagenet for 2k clean examples for training. Best methods indicated in \textbf{bold face}.}
\label{tab:mini_imagenet}
\resizebox{11cm}{!}{ 
\begin{tabular}{l|l|l|c|c|c|c|c|c}
     &             &  & \multicolumn{6}{c}{Attack} \\   
Dataset & Architecture & Detector & FGSM &  PGD &  JSMA & EAD & CWL2 &  Noise \\
\hline
\multirow{3}{*}{mini-ImageNet} & \multirow{3}{*}{ResNet18} & LID-OC &   0.533 &  0.537 &   0.502 &  0.524 &   0.528 &    0.498 \\
     &          & DeepMD-OC &   0.449 &  0.472 &   0.476 &  0.470 &   0.464 &    0.558 \\
     &          & DeepRP (ours) &   \textbf{0.790} &  \textbf{0.760} &   \textbf{0.749} &  \textbf{0.744} &   \textbf{0.760} &    \textbf{0.871} \\
\hline
\end{tabular}
}
\end{table*}

\begin{table*}[t!]
\centering
\caption{AUC score for detection of white-box targeted perturbation attacks on CIFAR10. Best methods indicated in \textbf{bold face}.}
\label{tab:targeted}
\resizebox{10cm}{!}{ 
\begin{tabular}{l|l|l|c|c|c|c}
     &             &  & \multicolumn{4}{c}{Attack} \\   
Dataset & Architecture & Detector & FGSM &  PGD &  JSMA & CWL2 \\
\hline
\multirow{6}{*}{CIFAR10} & \multirow{3}{*}{ResNet18} & LID-OC &   0.365 &  0.475 &   0.525 &   0.513 \\
        &          & DeepMD-OC &   0.500 &  0.618 &   0.655 &   0.614 \\
        &          & DeepRP &   \textbf{0.517} &  \textbf{0.858} &   \textbf{0.872} &   \textbf{0.878} \\
\cline{2-7}
        & \multirow{3}{*}{ResNet34} & LID-OC &   \textbf{0.532} &  0.511 &   0.497 &   0.515 \\
        &          & DeepMD-OC &   0.500 &  0.563 &   0.571 &   0.577 \\
        &          & DeepRP &   0.528 &  \textbf{0.870} &   \textbf{0.885} &   \textbf{0.871} \\
\hline
\end{tabular}
}
\end{table*}

\cleardoublepage
\clearpage

{\small
\bibliographystyle{ieee_fullname}
\bibliography{egbib}

\begin{thebibliography}{10}\itemsep=-1pt

\bibitem{abdelzaher2018toward}
Tarek Abdelzaher, Nora Ayanian, Tamer Basar, Suhas Diggavi, Jana Diesner,
  Deepak Ganesan, Ramesh Govindan, Susmit Jha, Tancrede Lepoint, Benjamin
  Marlin, et~al.
\newblock Toward an internet of battlefield things: A resilience perspective.
\newblock {\em Computer}, 51(11):24--36, 2018.

\bibitem{bagnall2017training}
Alexander Bagnall, Razvan Bunescu, and Gordon Stewart.
\newblock Training ensembles to detect adversarial examples.
\newblock {\em arXiv preprint arXiv:1712.04006}, 2017.

\bibitem{bingham2001random}
Ella Bingham and Heikki Mannila.
\newblock Random projection in dimensionality reduction: applications to image
  and text data.
\newblock In {\em Proceedings of the seventh ACM SIGKDD international
  conference on Knowledge discovery and data mining}, pages 245--250, 2001.

\bibitem{boopathy2019cnn}
Akhilan Boopathy, Tsui-Wei Weng, Pin-Yu Chen, Sijia Liu, and Luca Daniel.
\newblock Cnn-cert: An efficient framework for certifying robustness of
  convolutional neural networks.
\newblock In {\em Proceedings of the AAAI Conference on Artificial
  Intelligence}, volume~33, pages 3240--3247, 2019.

\bibitem{bulusu2020anomalous}
Saikiran Bulusu, Bhavya Kailkhura, Bo Li, Pramod~K Varshney, and Dawn Song.
\newblock Anomalous example detection in deep learning: A survey.
\newblock {\em IEEE Access}, 8:132330--132347, 2020.

\bibitem{carlini2017adversarial}
Nicholas Carlini and David Wagner.
\newblock Adversarial examples are not easily detected: Bypassing ten detection
  methods.
\newblock In {\em Proceedings of the 10th ACM Workshop on Artificial
  Intelligence and Security}, pages 3--14, 2017.

\bibitem{carlini2017towards}
Nicholas Carlini and David Wagner.
\newblock Towards evaluating the robustness of neural networks.
\newblock In {\em 2017 ieee symposium on security and privacy (sp)}, pages
  39--57. IEEE, 2017.

\bibitem{chen2017ead}
Pin-Yu Chen, Yash Sharma, Huan Zhang, Jinfeng Yi, and Cho-Jui Hsieh.
\newblock Ead: elastic-net attacks to deep neural networks via adversarial
  examples.
\newblock {\em arXiv preprint arXiv:1709.04114}, 2017.

\bibitem{crecchi2019detecting}
F Crecchi, D Bacciu, and B Biggio.
\newblock Detecting adversarial examples through nonlinear dimensionality
  reduction.
\newblock In {\em 27th European Symposium on Artificial Neural Networks,
  Computational Intelligence and Machine Learning, ESANN 2019}, pages 483--488.
  ESANN (i6doc. com), 2019.

\bibitem{dasgupta2013experiments}
Sanjoy Dasgupta.
\newblock Experiments with random projection.
\newblock {\em arXiv preprint arXiv:1301.3849}, 2013.

\bibitem{dasgupta1999elementary}
Sanjoy Dasgupta and Anupam Gupta.
\newblock An elementary proof of the johnson-lindenstrauss lemma.
\newblock {\em International Computer Science Institute, Technical Report},
  22(1):1--5, 1999.

\bibitem{feinman2017detecting}
Reuben Feinman, Ryan~R Curtin, Saurabh Shintre, and Andrew~B Gardner.
\newblock Detecting adversarial samples from artifacts.
\newblock {\em arXiv preprint arXiv:1703.00410}, 2017.

\bibitem{fidel2019explainability}
Gil Fidel, Ron Bitton, and Asaf Shabtai.
\newblock When explainability meets adversarial learning: Detecting adversarial
  examples using shap signatures.
\newblock {\em arXiv preprint arXiv:1909.03418}, 2019.

\bibitem{fort2021exploring}
Stanislav Fort, Jie Ren, and Balaji Lakshminarayanan.
\newblock Exploring the limits of out-of-distribution detection.
\newblock {\em arXiv preprint arXiv:2106.03004}, 2021.

\bibitem{gehr2018ai2}
Timon Gehr, Matthew Mirman, Dana Drachsler-Cohen, Petar Tsankov, Swarat
  Chaudhuri, and Martin Vechev.
\newblock Ai2: Safety and robustness certification of neural networks with
  abstract interpretation.
\newblock In {\em 2018 IEEE Symposium on Security and Privacy (SP)}, pages
  3--18. IEEE, 2018.

\bibitem{goodfellow2014explaining}
Ian~J Goodfellow, Jonathon Shlens, and Christian Szegedy.
\newblock Explaining and harnessing adversarial examples.
\newblock {\em arXiv preprint arXiv:1412.6572}, 2014.

\bibitem{grosse2017statistical}
Kathrin Grosse, Praveen Manoharan, Nicolas Papernot, Michael Backes, and
  Patrick McDaniel.
\newblock On the (statistical) detection of adversarial examples.
\newblock {\em arXiv preprint arXiv:1702.06280}, 2017.

\bibitem{he2016deep}
Kaiming He, Xiangyu Zhang, Shaoqing Ren, and Jian Sun.
\newblock Deep residual learning for image recognition.
\newblock In {\em Proceedings of the IEEE conference on computer vision and
  pattern recognition}, pages 770--778, 2016.

\bibitem{hecht1994context}
Robert Hecht-Nielsen et~al.
\newblock Context vectors: general purpose approximate meaning representations
  self-organized from raw data.
\newblock {\em Computational intelligence: Imitating life}, 3(11):43--56, 1994.

\bibitem{hsu2020generalized}
Yen-Chang Hsu, Yilin Shen, Hongxia Jin, and Zsolt Kira.
\newblock Generalized odin: Detecting out-of-distribution image without
  learning from out-of-distribution data.
\newblock In {\em Proceedings of the IEEE/CVF Conference on Computer Vision and
  Pattern Recognition}, pages 10951--10960, 2020.

\bibitem{huang2017densely}
Gao Huang, Zhuang Liu, Laurens Van Der~Maaten, and Kilian~Q Weinberger.
\newblock Densely connected convolutional networks.
\newblock In {\em Proceedings of the IEEE conference on computer vision and
  pattern recognition}, pages 4700--4708, 2017.

\bibitem{ilyas2019adversarial}
Andrew Ilyas, Shibani Santurkar, Dimitris Tsipras, Logan Engstrom, Brandon
  Tran, and Aleksander Madry.
\newblock Adversarial examples are not bugs, they are features.
\newblock In {\em Advances in Neural Information Processing Systems}, pages
  125--136, 2019.

\bibitem{johnson1984extensions}
William~B Johnson and Joram Lindenstrauss.
\newblock Extensions of lipschitz mappings into a hilbert space.
\newblock {\em Contemporary mathematics}, 26(189-206):1, 1984.

\bibitem{ko2021unsupervised}
Gihyuk Ko and Gyumin Lim.
\newblock Unsupervised detection of adversarial examples with model
  explanations.
\newblock {\em arXiv preprint arXiv:2107.10480}, 2021.

\bibitem{koner2021oodformer}
Rajat Koner, Poulami Sinhamahapatra, Karsten Roscher, Stephan G{\"u}nnemann,
  and Volker Tresp.
\newblock Oodformer: Out-of-distribution detection transformer.
\newblock {\em arXiv preprint arXiv:2107.08976}, 2021.

\bibitem{krahmer2011new}
Felix Krahmer and Rachel Ward.
\newblock New and improved johnson--lindenstrauss embeddings via the restricted
  isometry property.
\newblock {\em SIAM Journal on Mathematical Analysis}, 43(3):1269--1281, 2011.

\bibitem{kumar2020curse}
Aounon Kumar, Alexander Levine, Tom Goldstein, and Soheil Feizi.
\newblock Curse of dimensionality on randomized smoothing for certifiable
  robustness.
\newblock {\em arXiv preprint arXiv:2002.03239}, 2020.

\bibitem{lecuyer2019certified}
Mathias Lecuyer, Vaggelis Atlidakis, Roxana Geambasu, Daniel Hsu, and Suman
  Jana.
\newblock Certified robustness to adversarial examples with differential
  privacy.
\newblock In {\em 2019 IEEE Symposium on Security and Privacy (SP)}, pages
  656--672. IEEE, 2019.

\bibitem{lee2018simple}
Kimin Lee, Kibok Lee, Honglak Lee, and Jinwoo Shin.
\newblock A simple unified framework for detecting out-of-distribution samples
  and adversarial attacks.
\newblock In {\em Advances in Neural Information Processing Systems}, pages
  7167--7177, 2018.

\bibitem{levine2019wasserstein}
Alexander Levine and Soheil Feizi.
\newblock Wasserstein smoothing: Certified robustness against wasserstein
  adversarial attacks.
\newblock {\em arXiv preprint arXiv:1910.10783}, 2019.

\bibitem{li2017adversarial}
Xin Li and Fuxin Li.
\newblock Adversarial examples detection in deep networks with convolutional
  filter statistics.
\newblock In {\em Proceedings of the IEEE International Conference on Computer
  Vision}, pages 5764--5772, 2017.

\bibitem{liang2018detecting}
Bin Liang, Hongcheng Li, Miaoqiang Su, Xirong Li, Wenchang Shi, and Xiaofeng
  Wang.
\newblock Detecting adversarial image examples in deep neural networks with
  adaptive noise reduction.
\newblock {\em IEEE Transactions on Dependable and Secure Computing}, 2018.

\bibitem{lyu2019fastened}
Zhaoyang Lyu, Ching-Yun Ko, Zhifeng Kong, Ngai Wong, Dahua Lin, and Luca
  Daniel.
\newblock Fastened crown: Tightened neural network robustness certificates.
\newblock {\em arXiv preprint arXiv:1912.00574}, 2019.

\bibitem{ma2018deepgauge}
Lei Ma, Felix Juefei-Xu, Fuyuan Zhang, Jiyuan Sun, Minhui Xue, Bo Li, Chunyang
  Chen, Ting Su, Li Li, Yang Liu, et~al.
\newblock Deepgauge: Multi-granularity testing criteria for deep learning
  systems.
\newblock In {\em Proceedings of the 33rd ACM/IEEE International Conference on
  Automated Software Engineering}, pages 120--131, 2018.

\bibitem{ma2018combinatorial}
Lei Ma, Fuyuan Zhang, Minhui Xue, Bo Li, Yang Liu, Jianjun Zhao, and Yadong
  Wang.
\newblock Combinatorial testing for deep learning systems.
\newblock {\em arXiv preprint arXiv:1806.07723}, 2018.

\bibitem{ma2018characterizing}
Xingjun Ma, Bo Li, Yisen Wang, Sarah~M Erfani, Sudanthi Wijewickrema, Grant
  Schoenebeck, Dawn Song, Michael~E Houle, and James Bailey.
\newblock Characterizing adversarial subspaces using local intrinsic
  dimensionality.
\newblock {\em arXiv preprint arXiv:1801.02613}, 2018.

\bibitem{madry2017towards}
Aleksander Madry, Aleksandar Makelov, Ludwig Schmidt, Dimitris Tsipras, and
  Adrian Vladu.
\newblock Towards deep learning models resistant to adversarial attacks.
\newblock {\em arXiv preprint arXiv:1706.06083}, 2017.

\bibitem{magen2002dimensionality}
Avner Magen.
\newblock Dimensionality reductions that preserve volumes and distance to
  affine spaces, and their algorithmic applications.
\newblock In {\em International Workshop on Randomization and Approximation
  Techniques in Computer Science}, pages 239--253. Springer, 2002.

\bibitem{metzen2017detecting}
Jan~Hendrik Metzen, Tim Genewein, Volker Fischer, and Bastian Bischoff.
\newblock On detecting adversarial perturbations.
\newblock {\em arXiv preprint arXiv:1702.04267}, 2017.

\bibitem{miller2017not}
David~J Miller, Yulia Wang, and George Kesidis.
\newblock When not to classify: Anomaly detection of attacks (ada) on dnn
  classifiers at test time.
\newblock {\em arXiv preprint arXiv:1712.06646}, 2017.

\bibitem{mohseni2020self}
Sina Mohseni, Mandar Pitale, JBS Yadawa, and Zhangyang Wang.
\newblock Self-supervised learning for generalizable out-of-distribution
  detection.
\newblock In {\em Proceedings of the AAAI Conference on Artificial
  Intelligence}, volume~34, pages 5216--5223, 2020.

\bibitem{ndiour2020out}
Ibrahima Ndiour, Nilesh Ahuja, and Omesh Tickoo.
\newblock Out-of-distribution detection with subspace techniques and
  probabilistic modeling of features.
\newblock {\em arXiv preprint arXiv:2012.04250}, 2020.

\bibitem{papernot2018deep}
Nicolas Papernot and Patrick McDaniel.
\newblock Deep k-nearest neighbors: Towards confident, interpretable and robust
  deep learning.
\newblock {\em arXiv preprint arXiv:1803.04765}, 2018.

\bibitem{papernot2016limitations}
Nicolas Papernot, Patrick McDaniel, Somesh Jha, Matt Fredrikson, Z~Berkay
  Celik, and Ananthram Swami.
\newblock The limitations of deep learning in adversarial settings.
\newblock In {\em 2016 IEEE European symposium on security and privacy
  (EuroS\&P)}, pages 372--387. IEEE, 2016.

\bibitem{pei2017deepxplore}
Kexin Pei, Yinzhi Cao, Junfeng Yang, and Suman Jana.
\newblock Deepxplore: Automated whitebox testing of deep learning systems.
\newblock In {\em proceedings of the 26th Symposium on Operating Systems
  Principles}, pages 1--18, 2017.

\bibitem{quintanilha2018detecting}
Igor~M Quintanilha, Roberto de ME~Filho, Jos{\'e} Lezama, Mauricio Delbracio,
  and Leonardo~O Nunes.
\newblock Detecting out-of-distribution samples using low-order deep features
  statistics.
\newblock 2018.

\bibitem{raghunathan2018certified}
Aditi Raghunathan, Jacob Steinhardt, and Percy Liang.
\newblock Certified defenses against adversarial examples.
\newblock {\em arXiv preprint arXiv:1801.09344}, 2018.

\bibitem{rauber2017foolbox}
Jonas Rauber, Wieland Brendel, and Matthias Bethge.
\newblock Foolbox: A python toolbox to benchmark the robustness of machine
  learning models.
\newblock {\em arXiv preprint arXiv:1707.04131}, 2017.

\bibitem{roth2019odds}
Kevin Roth, Yannic Kilcher, and Thomas Hofmann.
\newblock The odds are odd: A statistical test for detecting adversarial
  examples.
\newblock In {\em International Conference on Machine Learning}, pages
  5498--5507, 2019.

\bibitem{scholkopf2001estimating}
Bernhard Sch{\"o}lkopf, John~C Platt, John Shawe-Taylor, Alex~J Smola, and
  Robert~C Williamson.
\newblock Estimating the support of a high-dimensional distribution.
\newblock {\em Neural computation}, 13(7):1443--1471, 2001.

\bibitem{singla2019robustness}
Sahil Singla and Soheil Feizi.
\newblock Robustness certificates against adversarial examples for relu
  networks.
\newblock {\em arXiv preprint arXiv:1902.01235}, 2019.

\bibitem{sun2018testing}
Youcheng Sun, Xiaowei Huang, Daniel Kroening, James Sharp, Matthew Hill, and
  Rob Ashmore.
\newblock Testing deep neural networks.
\newblock {\em arXiv preprint arXiv:1803.04792}, 2018.

\bibitem{sun2018concolic}
Youcheng Sun, Min Wu, Wenjie Ruan, Xiaowei Huang, Marta Kwiatkowska, and Daniel
  Kroening.
\newblock Concolic testing for deep neural networks.
\newblock In {\em Proceedings of the 33rd ACM/IEEE International Conference on
  Automated Software Engineering}, pages 109--119, 2018.

\bibitem{tian2018detecting}
Shixin Tian, Guolei Yang, and Ying Cai.
\newblock Detecting adversarial examples through image transformation.
\newblock In {\em Thirty-Second AAAI Conference on Artificial Intelligence},
  2018.

\bibitem{wang2019evaluating}
Lu Wang, Xuanqing Liu, Jinfeng Yi, Zhi-Hua Zhou, and Cho-Jui Hsieh.
\newblock Evaluating the robustness of nearest neighbor classifiers: A
  primal-dual perspective.
\newblock {\em arXiv preprint arXiv:1906.03972}, 2019.

\bibitem{weng2019proven}
Lily Weng, Pin-Yu Chen, Lam Nguyen, Mark Squillante, Akhilan Boopathy, Ivan
  Oseledets, and Luca Daniel.
\newblock Proven: Verifying robustness of neural networks with a probabilistic
  approach.
\newblock In {\em International Conference on Machine Learning}, pages
  6727--6736, 2019.

\bibitem{weng2018towards}
Lily Weng, Huan Zhang, Hongge Chen, Zhao Song, Cho-Jui Hsieh, Luca Daniel,
  Duane Boning, and Inderjit Dhillon.
\newblock Towards fast computation of certified robustness for relu networks.
\newblock In {\em International Conference on Machine Learning}, pages
  5276--5285, 2018.

\bibitem{wong2018provable}
Eric Wong and Zico Kolter.
\newblock Provable defenses against adversarial examples via the convex outer
  adversarial polytope.
\newblock In {\em International Conference on Machine Learning}, pages
  5286--5295, 2018.

\bibitem{xu2017feature}
Weilin Xu, David Evans, and Yanjun Qi.
\newblock Feature squeezing: Detecting adversarial examples in deep neural
  networks.
\newblock {\em arXiv preprint arXiv:1704.01155}, 2017.

\end{thebibliography}
}

\end{document}